# Programmable Synchronization Graphs for Adaptive and Fault-Tolerant Modular Miniature Robots

O. Kulekcioglu,[1] A. B. Ahmad,[1] I. Garcia,[2] F. S. Alves,[2] O. Ozcan,[1] M. S. Hanay[1,2]*

[1]Department of Mechanical Engineering, Bilkent University; Ankara, Türkiye.

[2]International Iberian Nanotechnology Laboratory (INL); Braga, Portugal.

*Corresponding author. Email: selim.hanay@inl.int

**Abstract:** Modular miniature robots could provide scalable function in constrained environments, but coordinating many imperfect modules remains difficult when computation, communication and reliability are limited. A central robotics challenge is to coordinate many actuator-sensor modules without assigning a privileged leader, prescribing a fixed gait template, or relying on dense communication. Here we introduce a programmable synchronization-graph framework for modular miniature robots in which each actuator-sensor pair is represented as a network node and locomotor coordination is encoded through graph coupling. Fixed intra-subgraph links synchronize heterogeneous actuator groups, whereas a small number of signed inter-subgraph links program phase relationships between groups. In physical robot collectives with up to nine modules, graph coupling drives the emergence of synchronization, signed links tune the phase difference from in-phase to out-of-phase motion, and floor experiments produce gallop-like and trot-like contact patterns in a five-module robot assembly. Replacing dense all-to-all coupling with sparse d-regular topologies preserves synchronization while reducing the coupling burden. The same graph representation also captures fault tolerance: increasing graph degree increases the number of module deactivations tolerated before desynchronization. Finally, an upper-confidence-bound edge-selection algorithm learns inter-subgraph links that drive the system toward target phase states. In a separate deactivation benchmark, the graph-based controller avoids the leader-specific failure mode observed in centralized leader-follower control and reduces worst-case phase error by about threefold. These results establish programmable network topology as a compact control layer for gait phase programming, online adaptation and robustness to unit loss in modular miniature robots.

## INTRODUCTION

Miniature robots are increasingly being developed for operation in constrained, cluttered and communication-degraded environments, including inspection, search-and-rescue, environmental monitoring and security applications.[1-10] As these systems shrink in size while gaining functional complexity, conventional control architectures face a scaling problem: centralized computation, high-bandwidth communication and leader-follower coordination become difficult to maintain under tight limits in power, cost, volume and reliability.[11-12] For modular miniature robots, this scaling problem becomes a *coordination problem*: the controller must transform many imperfect actuator-sensor modules into a coherent locomotor system despite actuator variability, sensor drift, limited communication, and occasional loss of individual modules.

One response is to distribute capability across physically coupled modules. Modular and self-reconfigurable robotic systems can provide versatility, scalability and fault tolerance while enabling system-level functions that are difficult for a single miniature unit to achieve alone.[13-14] Recent small-scale collective and reconfigurable robots have further shown that changing morphology and inter-unit connectivity can adapt locomotion and task performance.[15] Once capability is distributed, however, the main design challenge shifts from the construction of each module to the coordination, synchronization and planning rules that allow the collective to act as a coherent machine under limited communication and connection strength.[16]

Embodied and physical computing provide an appealing route to this problem because computation can be partly carried by the dynamics of the robot body and its interactions rather than by a large, centralized processor. A recent perspective on physical computing in soft robots emphasizes that useful embodied controllers require not only input encoding and output decoding, but also a programmable kernel whose input-output evolution can be configured for different tasks.[17] Recent work has also demonstrated that nonlinear physical dynamics can serve as computational substrates in hardware, for example through reservoir-computing architectures based on nanomechanical resonators, further illustrating the potential of exploiting physical dynamics rather than conventional digital computation for intelligent systems.[18] Oscillator-based controllers, including central pattern generators and analog oscillator circuits, are particularly powerful for rhythmic locomotion because they generate coordinated motion with simple, robust dynamics.[3, 17] Yet analog oscillator systems face a central limitation: they suffer from programmability and scalability constraints, because frequency and phase relationships are often hard-wired by geometry, mechanical architecture and material properties; re-programming can require redesign rather than parameter tuning, and fabrication tolerances can introduce oscillator mismatches that degrade synchronization as the system grows.[17]

This creates a gap for modular miniature robotics. The desired controller should retain the robustness and natural coordination of oscillator networks, but it should also allow phase relationships, gait modes, network connectivity and recovery behavior to be programmed or learned after fabrication. For robotic use, the graph should not only describe which modules communicate; it should act as a control interface. Changing a small number of edges should switch gait phase, reducing edges should lower communication and computation burden, and learning edges online should compensate for hardware variability or unit loss. Centralized policies can impose such coordination, but they are vulnerable to single points of failure and often assign fixed roles such as leader and follower modules.[1, 7, 19-20] Decentralized approaches such as central pattern generators (CPGs) and local behavioural rules reduce this fragility[3, 6, 21-22] but they do not by themselves provide a general, topology-level way to program sparse coupling, switch gaits and adapt links in response to performance. What is missing is an experimental framework in which the graph itself —its connectivity, edge signs and adaptive link-selection rule— becomes the programmable control object.

Here we introduce a programmable synchronization-graph framework for modular miniature robots (Fig. 1). In this framework, each actuator-sensor pair is treated as a sensorimotor node, and coordination is encoded in the synchronization state of a graph. Fixed intra-subgraph edges establish coherent motion within groups of actuators, while a small number of signed, dynamic inter-subgraph edges set the phase relationship between groups and therefore program gait-like motion. This, gait selection is shifter from preassigned module identities or handcrafted timing patterns to graph topology. The same edge variables can also be selected online by a learning algorithm, turning network topology into a control layer rather than a static description of hardware. The main contribution is an experimental demonstration that synchronization-graph topology can be programmed, sparsified and adapted as a physical control layer for modular robot coordination under communication and failure constraints.

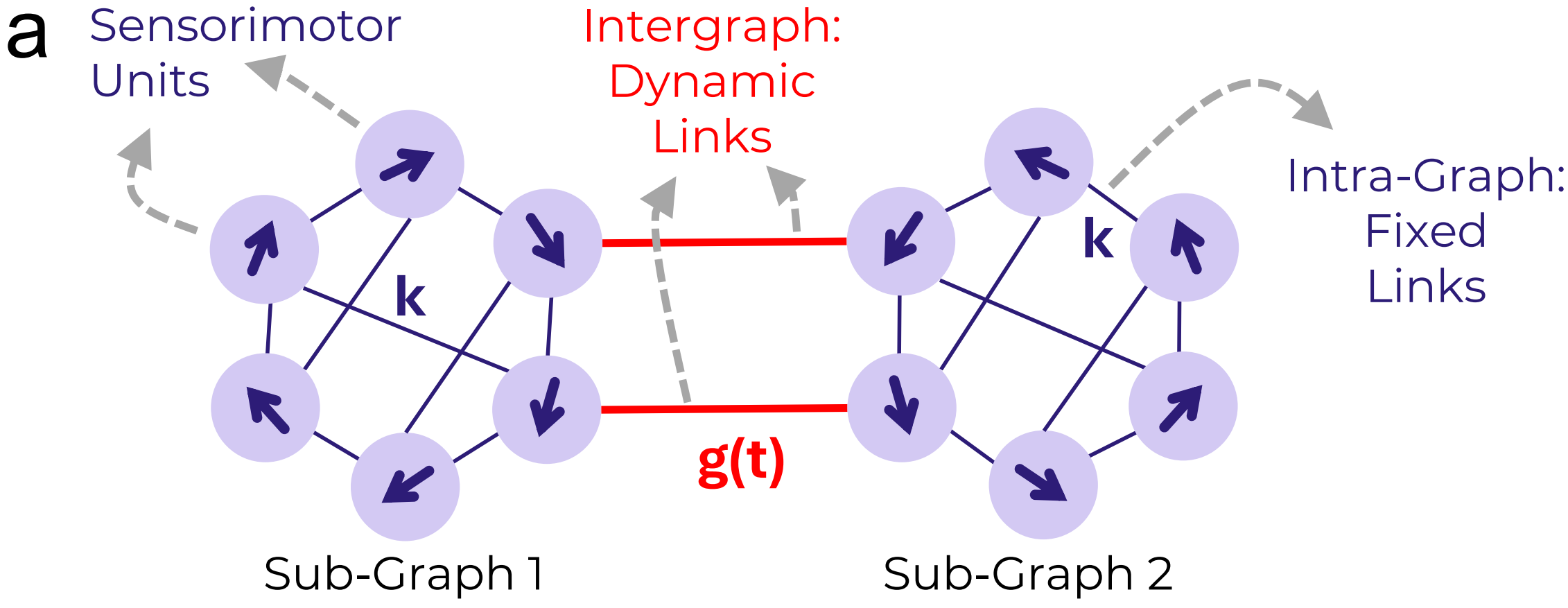

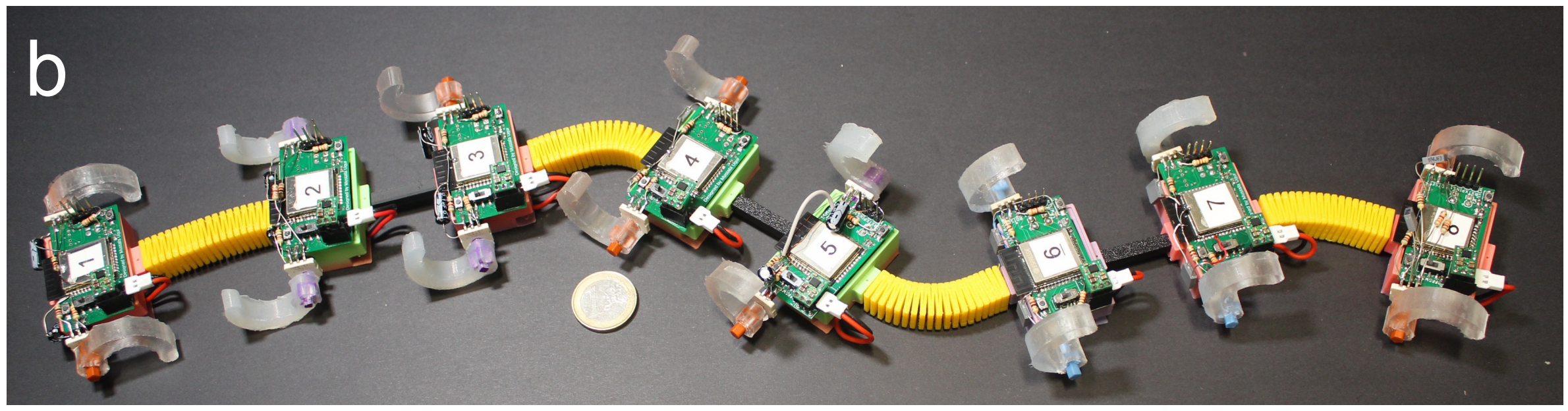

**Fig. 1. Modular Robots as an Adaptable Graph** (**A**) Conceptual representation of a robot as a network of sensorimotor units. Here each node (vertex) represents a sensorimotor unit, an actuator and its corresponding sensor reporting the instantaneous angle. The angle information from each unit is then passed onto other nodes through the links (edges) for computing the next value of the phase. Here, the graph is organized as two-subgraphs, corresponding to the left- and right-side actuators. The symbol ***k*** denotes the fixed intra-graph links, and as ***g(t)*** the dynamic inter-graph links. (**B**) General concept of the modular robotic assembly, illustrating an 8-unit system connected by an alternating sequence of elastic (yellow) and stiff (black) links. The selection of these links finetunes the inherent mechanical coupling. The 8-unit configuration depicted here is for illustrative purposes only to demonstrate the connectivity concept; the experimental on-floor scenarios evaluated in this work include 5-robot systems with varying attachment configurations.

The approach gives each module the same functional role and shifts the control variables from pre-assigned robot identities to graph couplings. In the present prototype, the graph-local update law is implemented through an ESP32 coordinator to allow controlled experiments across multiple robots; the coordinator computes timing updates and broadcasts commands for experimental synchronization, logging and repeatability, but it does not impose a leader trajectory or a fixed global gait template. Each update is defined by the phase differences along the graph edges, which is the same information that a distributed implementation would exchange between neighboring modules. Thus, the present system validates a graph-local control law using centralized experimental infrastructure, while avoiding the leader-follower structure tested later as a comparison. This makes it possible to test whether sparse programmable coupling can provide synchronization, gait phase control, robustness to unit loss and adaptive edge selection in a physical modular robot platform.

We demonstrated this framework using mechanically coupled modular robot units with up to nine modules. We first validated synchronization in heterogeneous actuator networks using all-to-all Kuramoto coupling within each subgraph.[23-25] We then showed that a small number of signed inter-subgraph links can control the phase difference between left and right actuator groups, producing in-phase, out-of-phase and intermediate gait-like coordination. To move beyond dense all-to-all coupling [26-27], we implemented sparse d-regular graph families and quantified how graph degree affects synchronization and robustness to random unit deactivation. Finally, we demonstrated online graph learning with a bandit edge-selection algorithm that forms inter-subgraph links until the desired phase relationship emerges. Together, these experiments position programmable network topology as a practical control substrate for resilient and adaptive modular miniature robots.

## RESULTS

### Programmable synchronization-graph architecture

We built the robot collective from mechanically coupled modular units (Fig. 1B). Each module contains two C-shaped actuators, one on the left and one on the right, and each actuator is paired with a rotary encoder that reports its instantaneous angle. In the graph representation, each actuator-sensor pair is one **sensorimotor node**; therefore, a platform with N robot modules contains 2N network nodes. The physical module design and soft mechanical links were described previously.[28]

The graph is partitioned into left and right subgraphs, corresponding to the actuators on the two sides of the platform (Fig. 1A). This partition reflects the locomotion requirement that actuators on the same side often need to coordinate, while the relative phase between sides determines gait-like motion. Within each subgraph, fixed couplings establish internal synchronization.

Only a small subset of links —the inter-subgraph links— is treated as programmable. Their sign and strength determine whether the two synchronized subgraphs attract, repel or settle at an intermediate phase. By separating fixed intra-subgraph synchronization from dynamic inter-subgraph coupling, the controller reduces the number of variables that must be adjusted while preserving a direct route to gait programming and online edge learning.

### Graph coupling synchronizes heterogeneous modules

We first asked whether graph coupling could synchronize heterogeneous physical actuators despite differences in free-running frequency. For this validation experiment, each left and right subgraph was initially implemented with all-to-all Kuramoto coupling and no inter-subgraph coupling (Fig. 2A). We tested networks containing N=7, 8 and 9 modules and quantified synchrony in each subgraph by using its order parameter r:

$$\boldsymbol{r} = r\, e^{i\psi} \equiv \frac{1}{N}\sum_{j=1}^{N} \mathrm{e}^{\theta_j} \quad \dots (1)$$

where the summation runs over the nodes in the specific subgraph and $\theta_j$ denotes the phase of the $j^{th}$ node. As the system synchronizes, the phase of each node converges to the same value, so that the magnitude of the order parameter approaches unity (r→1). Conversely, for a random distribution of N angles,[29] the order parameter scales as $1/\sqrt{N}$, approaching to zero in the limit of infinitely many random angles.

To achieve synchronization, different nodes interact with each other using the Kuramoto coupling term, so that the dynamics of the angle $\theta_i$ is expressed as:

$$\frac{d\theta_i}{dt} = \omega_i + \frac{k}{N}\sum_{j=1}^{N} \sin\left(\theta_j - \theta_i\right) \quad \dots (2)$$

Here $\omega_i$ is the uncoupled or free-running rotation frequency of the unit, $k$ is the intra-subgraph coupling strength. To emulate the variability expected in miniature robotic modules, each actuator was assigned a natural frequency drawn from a Gaussian distribution (mean Hz, standard deviation 0.2 Hz), and PWM values were calibrated so that each leg ran near its assigned frequency. The graph-local phase updates were computed on an ESP32 coordinator for experimental control and broadcast to the modules typically every 75 ms using ESP-NOW (time was varied between 40-100 ms for different runs).

During each sweep, the coupling strength k was increased from 0 to 200 in steps of 5. At each nonzero value of $k$, the network ran for 8 s to evaluate whether synchronization was reached by calculating the order parameter using equation (1). After each evaluation period, the coupling constant was reset to 0 for a duration of 0.4 s to clear history effects. The order parameter reported in Fig. 2A was calculated from steady-state windows after transients had decayed. These experiments were performed with the legs suspended in stationary holders to isolate electronic graph coupling from unintended mechanical synchronization through the robot bodies and soft links. An example of the network synchronization is shown in Supplementary Movie 1.

For each trial, the critical coupling was defined as the point at which the measured order parameter exceeded 0.90. This relatively high threshold was selected to investigate the relationship between the minimum coupling coefficient required for synchronization and the distribution of the intrinsic frequencies of the network. Across 10 trials for each network size, the critical coupling generally increased with the realized spread in natural frequencies (Supplementary Fig. S1) consistent with the expected behavior of oscillator networks.[30]

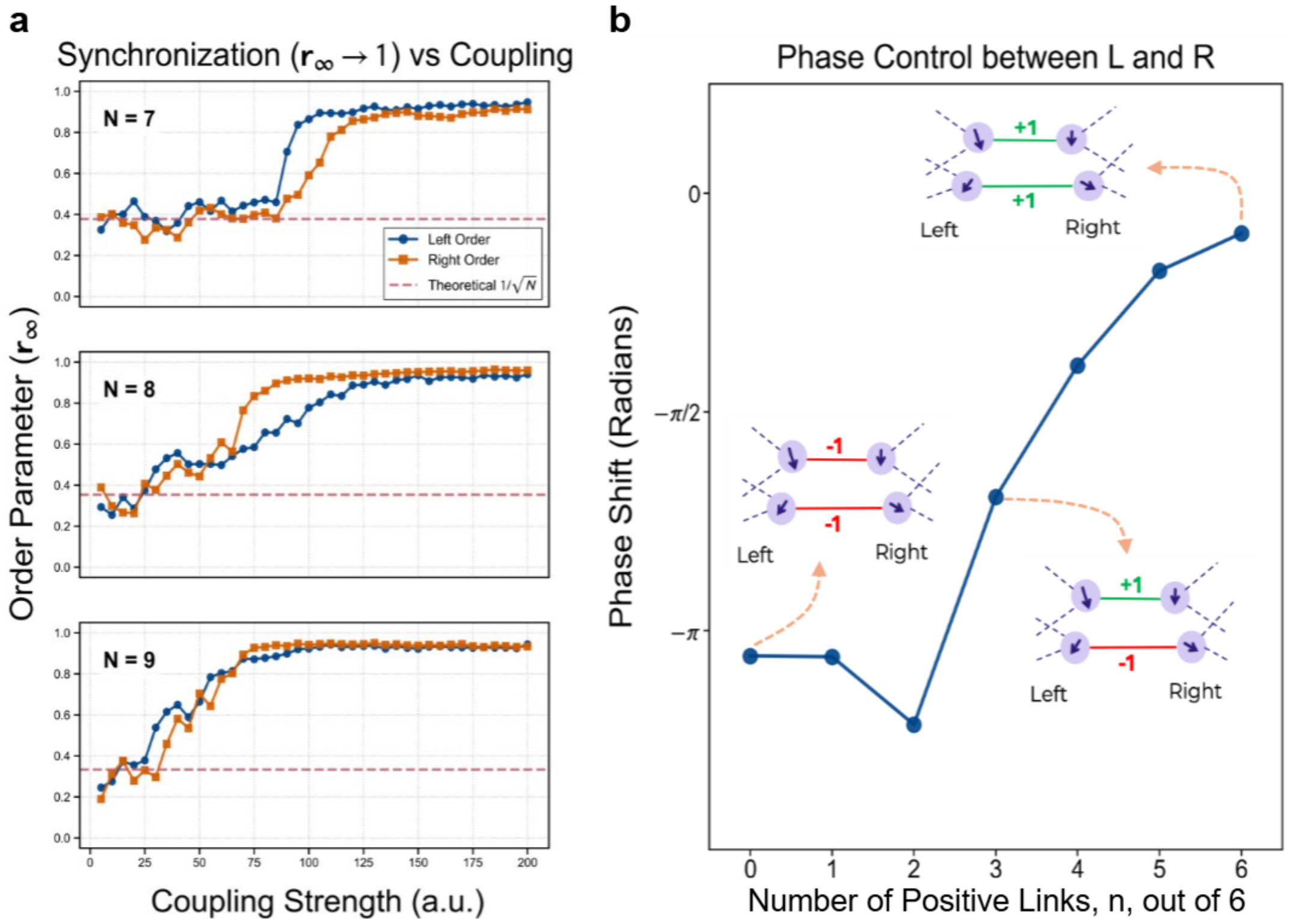

**Fig. 2. Programmable graph synchronization and phase control.** (**A**) The order parameter within the left and right subgraphs increases toward unity as the intra-subgraph coupling k is increased, showing synchronization of heterogeneous robot actuators for N=7, 8 and 9. The dashed line indicates the finite-size baseline for uncoupled phases. (**B**) The phase difference between left and right subgraphs is programmed by changing n, the number of positive signed inter-subgraph links while keeping the total number of links fixed at six. The plot is shown for N=9. All negative links (n=0) induce out-of-phase, trot-like coordination; all positive links (n=6) induce in-phase, gallop-like coordination; mixed signs produce intermediate phase offsets.

### Signed inter-subgraph links program gait phase

After establishing internal synchronization, we tested whether a small number of signed links between the left and right subgraphs could program the relative phase between the two actuator groups. In these experiments, the intra-subgraph couplings remained all-to-all, while six inter-subgraph links were introduced between predefined left-right pairs. With these inter-subgraph links, the phase dynamics of node *i* becomes:

$$\frac{d\theta_i}{dt} = \omega_i + \frac{k}{N}\sum_{j=1}^{N} \sin(\theta_j - \theta_i) \; + \frac{g}{N}\sum_{k \in \aleph(i)} \sin(\theta_k - \theta_i) \quad \ldots (3)$$

Here, $g$ is the inter-subgraph coupling strength and $\aleph(i)$ denotes the set of neighbors of node *i* in the opposite subgraph. We used the same interconnections and changed the links from all-negative to all-positive one by one.

The sign composition of these links controlled the phase relation between the two sub-graphs (Fig. 2B). When all inter subgraph links were positive, coupling drove the system towards in-phase motion; whereas all-negative coupling drove it toward an approximately $\pi$-shifted state. Mixtures of positive and negative links produced intermediate phase offsets, showing that gait phase can be programmed by a low-dimensional edge variable rather than by assigning different roles to individual robotic modules (Supplementary Movie 2).

We then tested whether this phase programming could alter physical locomotion on the floor using five coupled robot modules. The same motor-control and phase-update rules were used, while simple 3D-printed backbone elements of different stiffnesses prevented entanglement and limited uncontrolled mechanical synchronization (Supplementary Fig. S2). This experiment connects the graph-level phase state to observable gait-like contact pattern. With inter-subgraph coupling set positive, the left and right sides converged to in-phase, gallop-like coordination; with negative coupling, they converged to out-of-phase, trot-like coordination (Fig. 3 and Supplementary Movie 3). Although these demonstrations used dense inter-subgraph coupling, the later learning experiments showed that far fewer links can be sufficient to achieve target phase relationships.

**Sparse topology reduces the coupling burden**

All-to-all intra-subgraph coupling is useful for validating the synchronization law, but it scales poorly because the number of links grows quadratically with the number of modules. We therefore tested whether sparse d-regular graph topologies could preserve synchronization with fewer links. For N=8 robot modules, each left and right subgraph was initialized as a connected d-regular graph with d varying from 3 to 7 (with d=7 corresponding to all-to-all connectivity). For each d, ten random graphs (connected) were tested in stationary holders, and the steady-state order parameter was measured (Supplementary Fig. S3).

**Graph connectivity determines failure tolerance**

We next quantified how network connectivity affects robustness to module loss. In these experiments, randomly selected modules were deactivated so that they no longer sent or received synchronization updates, their last measured angles were used when computing updates for the remaining active nodes. This type of unit loss deliberately introduced persistent error bias and mimicked a communication or sensing failure rather than simply removing a node cleanly from the graph. The deactivated nodes continued to move with their inherent dynamics, not receiving any feedback signal from any other nodes.

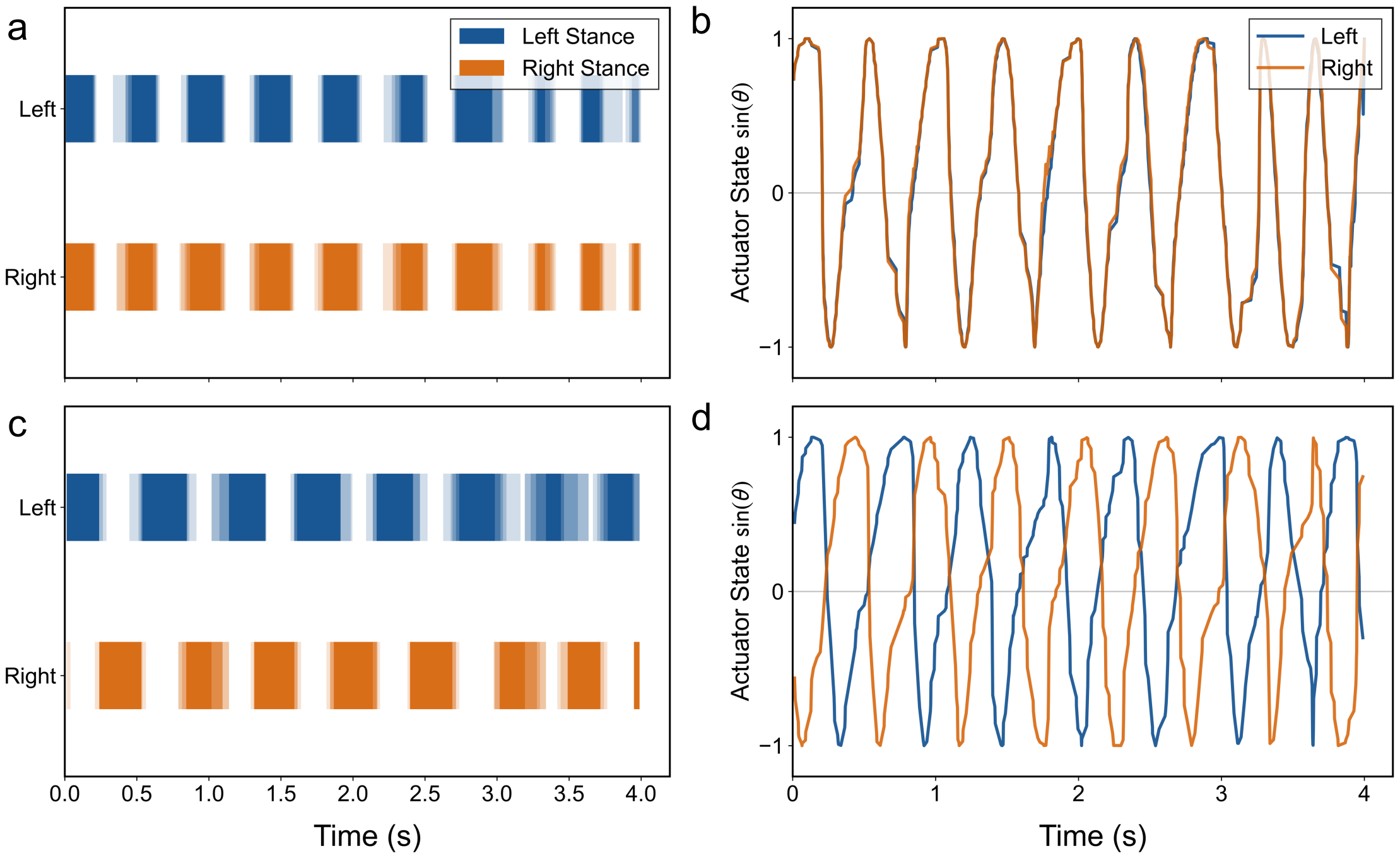

**Fig. 3. Floor locomotion under programmed phase relationships.** (**A**) Gait diagram for in-phase motion, where shading indicates the number of actuators touching the ground. (**B**) Locomotion phase data for in-phase motion, represented by the sine of actuator angle and averaged across the five left and right actuator groups. (**C, D**) Corresponding gait diagram and phase data for out-of-phase motion.

The dynamics of representative 3-regular network of N=8 is shown in Fig. 4A. The robot is synchronized in the beginning with all links intact, then links are removed one at a time. After the first two deactivations, the remaining active networks maintained high synchrony, but the order parameter dropped after a third deactivation at approximately 9.8 s, indicating loss of robust synchronization in at least one subgraph.

A representative 5-regular network tolerated a larger number of failures, maintaining synchronization until the fifth deactivation (Fig. 4B and Supplementary Movie 4.) The oscillations observed after desynchronization reflected partial synchronization, where most active units remained coordinated while one unit drifted with a constant frequency difference relative to the collective frequency of the rest of the group (Supplementary Fig. S4).

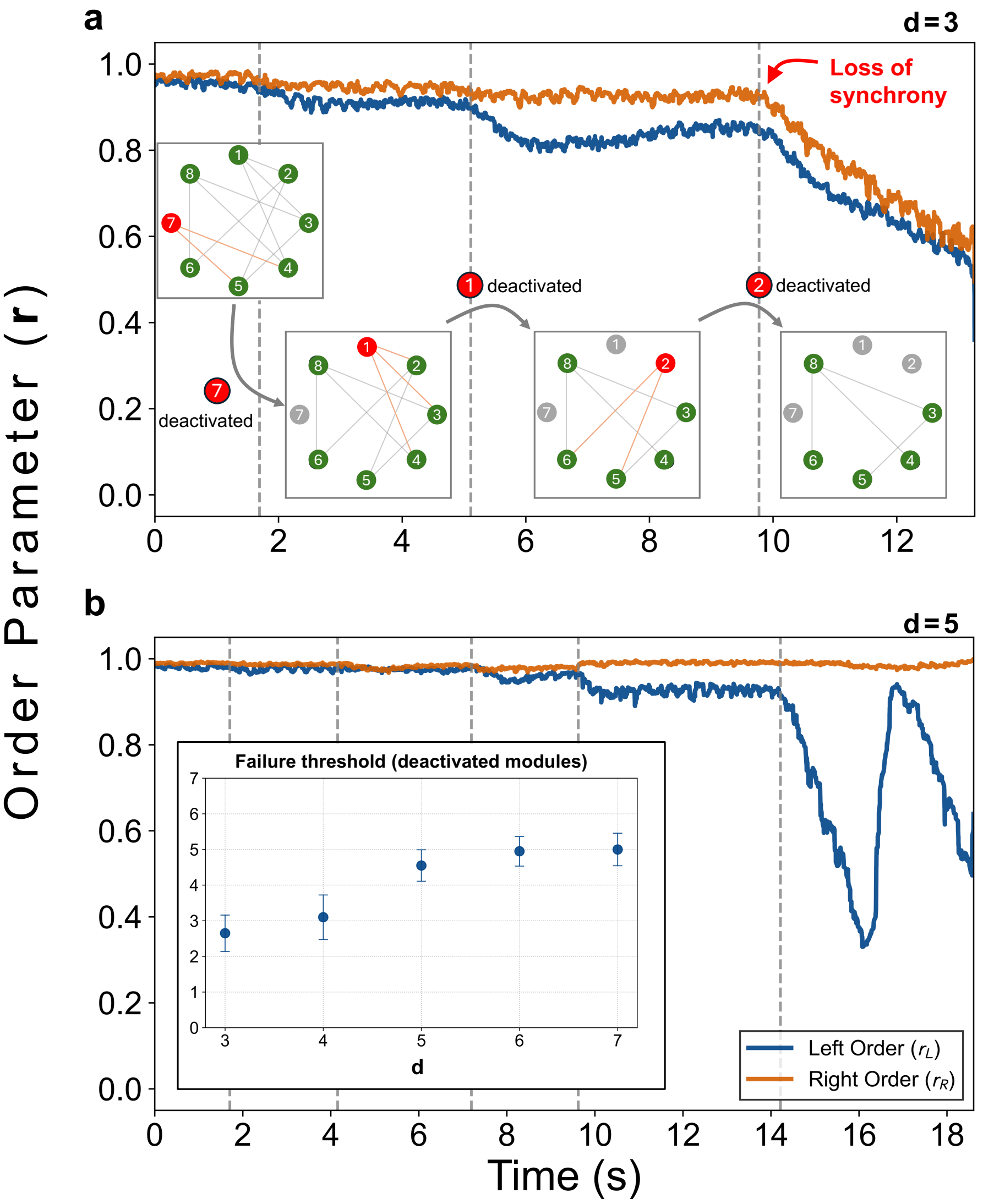

**Fig. 4. Robustness of a sparse robot network to unit loss.** (**A**) The evolution of the order parameter is shown as units lose their communication ability (both in the left and right subgraphs) at times indicated by dashed lines for the case of a 3-regular graph with N=8 robots. The loss of the first two random units (at t=1.7 s and t=5.1 s respectively) is tolerated by the rest of the network, but as the 3rd unit is disconnected around t= 9.8 s, both the Left and Right subgraphs lose their synchrony as evident by the decaying order parameters. Insets show the nodes and links lost in each step and the remaining network. (**B**) Robustness exhibited by a 5-regular network where units are lost one after another at t= 1.7, 4.2, 7.2, 9.6, and 14.2. The synchronization persisted until the loss of 5th unit at time t=14.2, at which point a single drifter emerged causing the observed behavior. **Inset:** Onset of failure as measured by the number of deactivations required for loss of synchronization as a function of the degree of the network (d) for d-regular networks. Error bars show 95% confidence intervals, calculated from a t-distribution with n-1 = 19.

To quantify robustness to unit loss, we measured the number of deactivated modules required to drive at least one subgraph below the synchronization threshold. For each degree ***d***, twenty trials were performed with independently generated random connected d-regular graphs. The system was classified as desynchronized when the order parameter, after being synchronized above $r > 0.8$, fell below $r = 0.6$ in at least one subgraph.

The failure threshold increased with graph degree (Inset of Fig. 4B). Networks with d=3 typically failed after two to three deactivations, whereas d=6 networks tolerated approximately five deactivations, approaching the robustness of the all-to-all d=7 case. This result shows that topology is not only an implementation detail: graph connectivity directly controls the fault tolerance of the modular robot collective.

**Online edge learning adapts inter-subgraph coupling**

We also tested whether the robot graph could learn to select inter-subgraph links online rather than relying on user-chosen edge signs. We implemented a contextual bandit edge-selection (CBE) algorithm using an upper confidence bound (UCB) rule. The algorithm evaluates candidate signed inter-subgraph links according to whether they move the measured phase difference (between the left and right subgraphs) toward the target value, retaining links that improve performance and removing those that do not.

For these tests, a five-module network was initialized with randomized intrinsic frequencies and internally synchronized left and right subgraphs by fixed intra-subgraph links. At the start of each learning episode, all inter-subgraph links were removed. The algorithm then selected among candidate links connecting corresponding left and right actuators, with each candidate allowed to be either positive or negative.

The reward was based on the reduction in average phase error relative to the target phase difference. After a candidate link was activated, the network ran a decision cycle with a period of 1.35 s, and the resulting phase difference was measured to update the reward function. If the candidate link resulted in a positive reward function, the link was kept in the graph; otherwise, it was removed. To avoid stopping on transient coincidences with the target phase, the algorithm required the phase error to remain within a prescribed tolerance for two consecutive decision cycles before declaring lock-in.

Fig. 5 shows a representative learning sequence. Each time the target value for the phase shift was changed (blue dashed lines), all the existing inter-subgraph links were cleared, and the algorithm searched for a new signed edge set. The phase difference converged toward the requested targets, while the net signed link count shifted positive for in-phase targets and negative for out-of-phase targets. Examples of the robots operating under CBE algorithm are shown in Supplementary Movie 5.

The changes in the number of active positive and negative links are plotted in Fig. 5B, and the net number of links (the difference between the number of positive and negative links) in Fig. 5C. The green markers show the times where the algorithm detects the locking-in to the target phase difference, i.e. the target phase is successfully attained.

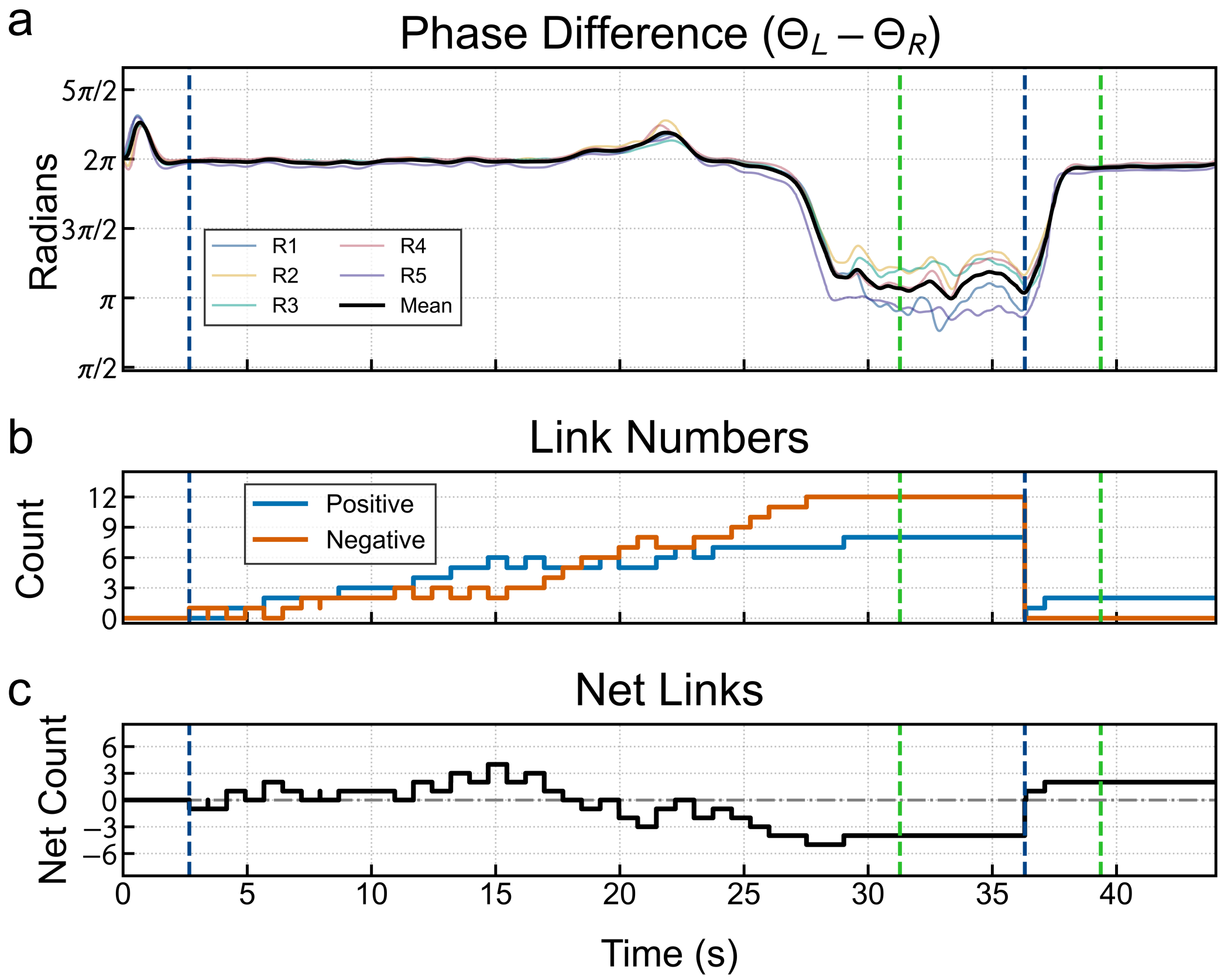

**Fig. 5. Graph learning algorithm CBE in action.** (**A**) The dynamics of the phase difference between the left and right subgraphs (thick curve) as well as those for individual robots. At each blue dashed line existing inter-graph links are terminated, a new target setpoint is assigned, and the algorithm starts to test and form new links. Green vertical dashed lines show when the system reaches the target phase values. The evolution of these links in terms of (**B**) positive and negative links, and (**C**) net number of links (positive – negative links) are shown.

For a given target value, the number of net links learned by the algorithm exhibited variances across trials since the robots were initialized with different intrinsic frequencies and individual modules possess small variations in their mechanical characteristics. Moreover, depending on the physical state of the modules, certain links cannot induce significant gait changes, resulting in negligible phase differences. During the activation of these links, the calculated reward might fluctuate around zero, potentially causing the algorithm to erroneously retain (or remove) unintended (intended) links. Still, the CBE algorithm demonstrates sufficient robustness to recover from these faulty link (de-)activations. Once a link is removed, a different link was tested within tens of milliseconds by the algorithm, which resulted in occasional spike-like features in the number of links on Fig. 5B,C. This variability is precisely where online edge selection is useful: the graph algorithm tunes itself to the existing hardware state in that instant, rather than assuming that a fixed link pattern will always produce the same phase value. Moreover, by avoiding the predefined universally positive or negative patterns, the algorithm identifies a reduced set of effective links. Eliminating these redundant links reduces the number of active coupling updates and may lower computational load in future onboard implementations.

**Deactivation Performance Comparison between Graph-Based and Centralized Control Algorithms**

Finally, to analyze the performance of the graph-based network under module communication loss, a comparison study was conducted. The objective of achieving synchronized *out-of-phase* motion for a network of *five robots*, was implemented by the CBE and leader-follower (LF) control algorithms on separate test runs and the results were compared. To minimize the effect of unintentional mechanical coupling between the robots, the five-robot set was divided into three sub-groups: robot #1 was mechanically disjointed, whereas the robots 2-3 and 4-5 were attached to each other with stiff backbones. To keep the essential control mechanism the same (Kuramoto update rule, equation 3) for each study, the leader-follower control algorithm was implemented using a directed graph structure. The leader robot was chosen as robot #1, and the rest of the robots were the followers. This was achieved by positively coupling the left leg of the leader to the left leg of each follower robot, and negatively coupling to each right leg, and vice versa for the right leg of the leader. However, only the inter-link between the left and right sides of the leader robot was undirected; all other links were directed in the sense that they were only affecting the follower robots and not interfering with the leader's actions. The coupling strength was kept the same for each connection. No other links were involved.

Before beginning the tests, the five-robot network was initialized with random frequencies as before, and these frequencies were kept the same for each trial (both for CBE and Leader-Follower) to reduce the stochastic effects on the comparison tests. For the deactivation tests with CBE algorithm, the same network structure was kept for consistency. During the tests, the robots from 1 to 5 were deactivated in turn. To deactivate a module, its control command to receive and transmit data was disabled, same as the deactivation tests conducted before. Once a unit was deactivated, it was kept disabled for ten seconds and turned back on. Another ten seconds were given for the system to recover back to the steady state at its target gait. This process went on until the 5$^{th}$ robot was disabled. The same deactivation order was followed for both control algorithms. Both control algorithms were tested ten times to increase the statistical accuracy of the results.

To evaluate the LF- and CBE-based networks, we defined a comparison metric based on *order parameters* which naturally measures synchronization related characteristics of a network. During the unit deactivation periods, the complex order parameters for the left ($\boldsymbol{r_L}$) and right ($\boldsymbol{r_R}$) legs were calculated for discrete segments of 50 ms duration. A difference vector, $\boldsymbol{\Delta r}[k] = \boldsymbol{r_L}[k] - \boldsymbol{r_R}[k]$, was computed for each segment $\boldsymbol{k}$, and its absolute value was subsequently time-averaged over the 10-second deactivation duration yielding 200 samples (with each sample 50 ms). The absolute value of this($m_i$) for the deactivation of *the* robot with index *i,* explicitly written out as:

$$m_i = \frac{1}{200}\sum_{k=1}^{200}|\boldsymbol{\Delta r}[k]| \quad \dots(4)$$

Theoretically, this metric converges to 2.0 in the ideal case of perfect intra-graph synchronization and an exact π inter-graph phase difference; and yields 0.0 for the perfect global synchrony case. Therefore, a reduction in this metric indicates a performance loss in achieving the target π phase difference within the physical robotic gait.

To compare the robustness of the graph-based CBE controller with that of a centralized leader-follower architecture, we implemented both controllers on the same five-robot platform and deactivated each robot in turn under matched conditions. Fig. 6A,B summarizes the resulting performance loss using the order-parameter-based metric defined in equation 4. In the CBE case, the error remains distributed more uniformly across the robots, consistent with the absence of a privileged leader. By contrast, the leader-follower controller exhibits a strong single-point vulnerability: deactivating the leader produces a much larger degradation than deactivating any follower. To complement this network-level metric, we also quantified the phase-error RMSE during deactivation (Fig. 6C, detailed in SI Section 2). This comparison showed that the graph-based CBE controller reduced the worst-case error from 1.88 ± 0.13 rad in the leader-follower case to 0.65 ± 0.22 rad in the CBE case, corresponding to an approximately threefold improvement in worst-case robustness; the average error was also lower for CBE (0.48 ± 0.07 rad) than for leader-follower control (0.64 ± 0.05 rad).

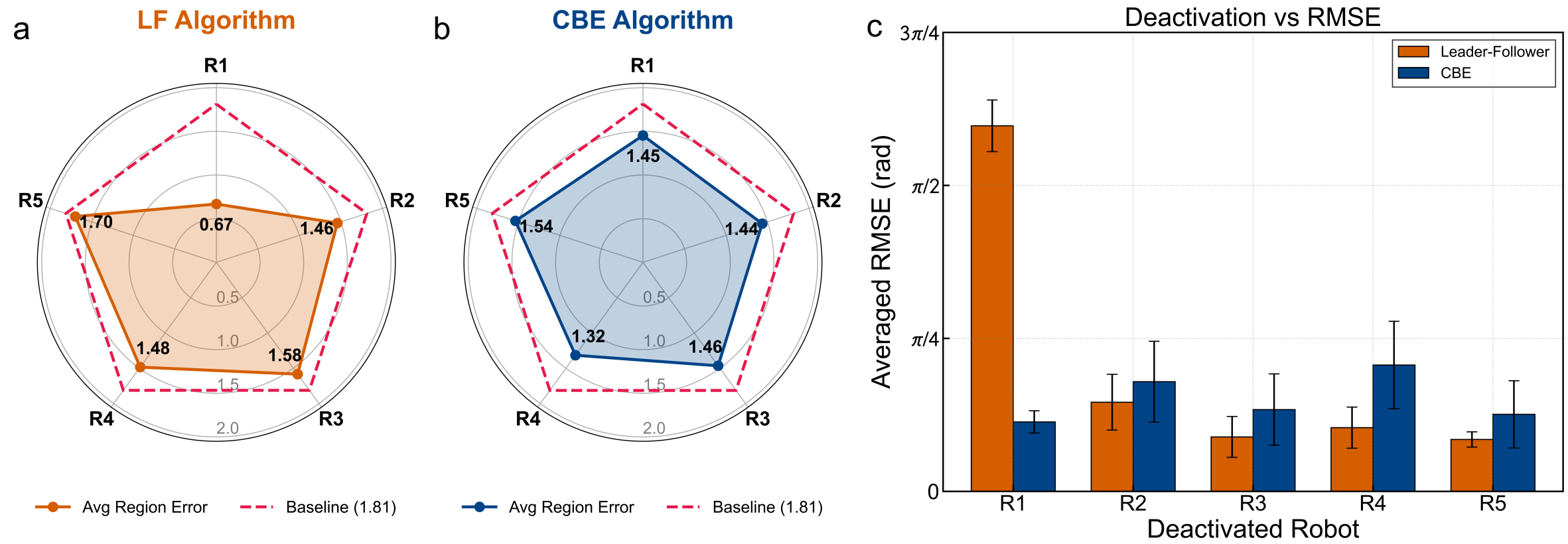

**Fig. 6. Performance comparison between centralized leader-follower and graph-based CBE control under unit deactivation.** (**A**) Star-diagram summary for the centralized leader-follower controller. The red dashed contour shows the baseline performance metric $m_i$ from steady-state runs with all units active, and the filled contour shows the mean value measured when the corresponding robot is deactivated. (**B**) Corresponding star-diagram summary for the graph-based CBE controller. The more uniform response across robots indicates that the graph-based CBE controller avoids the strong single-point vulnerability of the leader-follower architecture. (**C**) Mean phase-error RMSE during unit deactivation for the two controllers. The graph-based CBE controller reduces worst-case error and modestly improves average performance relative to leader-follower control.

## DISCUSSION

In this work, we introduced a programmable synchronization-graph framework for controlling modular miniature robots. By representing each actuator-sensor pair as a node and each coupling rule as an edge, the robot collective can be treated as a dynamical network whose synchronization state encodes coordinated motion. This formulation separates two control functions: fixed intra-subgraph links establish coherent motion within actuator groups, while signed inter-subgraph links program the phase relationship between groups and thereby select gait-like coordination modes. In this view, topology is not only a representation of the robot network; it is a control variable.

Using mechanically coupled robot modules, we showed that graph coupling can synchronize heterogeneous actuators, that a small number of signed links can tune the phase difference between left and right actuator groups, and that the same phase relationships produce distinct locomotion patterns on the floor. Moving beyond dense all-to-all coupling, sparse d-regular graph topologies preserved synchronization while reducing the number of required links. The robustness experiments further showed that graph connectivity directly affects failure tolerance: higher-degree networks maintained synchrony after a larger number of module deactivations. The CBE experiments demonstrated that inter-subgraph links can be learned online, allowing the robot graph to adapt its coupling structure toward target phase states rather than relying only on manually prescribed connections. Moreover, the graph-based CBE algorithm outperformed the centralized LF algorithm in both average and worst-case conditions.

The present study used an ESP32 coordinator to implement the graph-local update law and adopted controlled mechanical configurations to isolate the effect of programmable coupling from uncontrolled body-mediated synchronization. Accordingly, several experiments emphasized phase-based coordination and fault-tolerance metrics rather than task-level locomotion outcomes. These choices were made to isolate the core control question: whether programmable graph topology, rather than passive mechanical coupling or a handcrafted gait template, can determine phase relationships, coupling load, adaptive link selection and robustness to unit loss in a physical modular robot. The results should therefore be interpreted as a controlled demonstration of graph-programmable coordination rather than as a complete autonomous robotic system. Extending the framework to fully onboard distributed computation, larger robot collectives will convert this control layer into a more complete autonomous modular-robot architecture.

These results suggest that programmable network topology can serve as a practical control substrate for resilient modular robots. Unlike oscillator controllers whose phase relationships are fixed primarily by hardware design, the synchronization graph allows coordination patterns to be programmed, sparsified and adapted after fabrication. The present implementation uses a coordinator to compute graph-local updates for experimental control, but the same coupling rules could be distributed across modules in future systems. More broadly, this approach provides a route toward miniature robotic collectives in which robustness and adaptability emerge not from a centralized policy or a single leader module, but from the topology and dynamics of the robot network itself. The central design object becomes the synchronization graph: its connectivity, edge signs and adaptive-link selection rule determine how the collective coordinates and how robustly it responds to unit loss.

## MATERIALS AND METHODS

### Robot Control Details

The PWM signals are generated with the Motor Control PWM (MCPWM) for ESP32s. The source clock of ESP32 is 80 MHz, and the timer *resolution is 20 MHz (*each tick is 50ns), and it *counts upwards*. T*he maximum counter value (TOP) is 2000*, (so the total period is 100 us / 10 kHz frequency). The PWM is set by updating the comparator value, so the RHS of the update equations (equation 2 and 3) is set to be the new comparator value of the timer. It is clamped to 0 or 2000 when necessary. The PWM signal is generated by setting the drive pin HIGH when the timer is below the comparator, and LOW until the TOP value. Comparator value is *updated when the timer hits back to 0 (tez).* Recorded data was filtered to smooth out the transients. In Fig. 2A, the order parameters; Fig. 2B, the phase difference values were rolling averaged with a window of 10 samples before the final calculations. In Fig. 3, raw angle data traces are processed with 3rd order low-pass filter with a 6 Hz cutoff frequency. In Fig. 4 and S4, raw angle data traces are processed with 3rd order band-pass filter with a 1.7-3.5 Hz frequency band. After applying the filter, both the filtered angles and calculated order parameters were rolling averaged with a window of 10 samples. In Fig. 5A, phase difference data traces are processed with 3rd order low-pass filter with a 0.9 Hz cutoff frequency. Due to physical differences between the left and right leg, the phase difference data has a noise component at the common oscillation frequency (~0.8 Hz), which is the reason for such a low cutoff frequency. In Fig. 6, the phase difference data traces are processed with a 3rd order low-pass filter with 1.5 Hz cutoff frequency before the error calculations. In Fig. S3, the order parameters were rolling averaged with a window of 50 samples.

**Details of CBE Algorithm**

A complete (graph) network consisting of 5 robots was initialized with random intrinsic frequencies ($\omega_i$), similar to the situation depicted earlier in Fig. 2. A sufficiently large intra-coupling coefficient (***k***) is introduced, to have two internally synchronized subgraphs. At the beginning of the algorithm, there are no inter-graph links present. To synchronize the system into different phase states (i.e. in-phase and out-of-phase), the CBE algorithm tests the effect of formation of new links from pre-defined inter-link candidates and activates/deactivates them depending on whether they improve the performance or not. The candidate pairs for the CBE algorithm tests were randomly selected as 25 negative and 15 positive links for anti-phase target, and vice versa for in-phase target. As earlier, the links have a fixed strength **|*g*|** but they can be either positive or negative. As a result, the CBE algorithm has initially a total of 40 possible inter-graph link candidates to choose from.

For CBE algorithm to work, there is a need to define a “reward” for specific actions (link activations). Since the goal of this algorithm is to set the mean phase difference between left and right actuators of each robot to a target angle $\phi_t$ , the mean angles of the left and right legs ($\overline{\theta}_l$ and $\overline{\theta}_r$, respectively) are calculated in a circular averaging manner (i.e. $\overline{\theta_l} = \arctan 2(\sum_{i\in l}\sin(\theta_l), \sum_{i\in l}\cos(\theta_l))$, and similarly for the right network). Then, the mean phase difference is calculated by $\Delta\overline{\theta} = wrap\left(\overline{\theta_l} - \overline{\theta_r} - \phi_t\right)$, wrapped between 0 to $2\pi$. The goal of the algorithm is to ensure $\left|\Delta\overline{\theta}\right| < \varepsilon$, where $\varepsilon$ is the allowable error in radians, in this case chosen to be 0.50 radians.

When the CBE algorithm decides to activate an inter-link, instantaneous mean phase difference ($\Delta\overline{\theta}_{before}$) is calculated. Then, the system is run for 1.35 seconds, and the new phase difference ($\Delta\bar{\theta}_{after}$) is measured. The reward for the corresponding link is $r = |\Delta\bar{\theta}_{before}| - |\Delta\bar{\theta}_{after}|$.

The CBE algorithm new inter-link selection rule is as follows. If there is a candidate link not tried before, it is tried. After trying an inter-link $l$ , the upper confidence bound for it is computed as $UCB_l = \bar{r}_l + c\sqrt{\frac{\ln t}{N_l}}$, where $\bar{r}_l$ is the average reward for candidate link l, $t$ is the number of decisions so far, $N_l$ is the number of times the link $l$ is tried, $c$ is the exploration constant (set to be 0.5). For the new decision, the inter-link with the highest UCB value is activated. The links which yield a positive reward are kept in the network, whereas negative links are removed from it.

The CBE algorithm keeps running until the network locks into the target phase difference $\phi_t$ . Note that during the operation of the network, even without any inter-link, the phase difference could instantaneously coincide with the goal difference $\phi_t$ , which is not an actual locking-in, but a mere coincidence. To avoid the algorithm halting at such moments, a counter was implemented which waits for two decision cycles to check the phase difference. This way, momentary coincidences were not considered a termination signal for the algorithm. Activation or deactivation of the links was suspended as this counter is operational, in order not to interfere with the check procedure.

**Funding:** The authors acknowledge the support of the project with the reference n.º 1313551786, funded by FCT – Fundação para a Ciência e a Tecnologia, I.P. under the Portuguese Resilience and Recovery Plan, through the NextGenerationEU Fund.

**Author Contributions:** MSH, FSA, OO, and OK conceptualized the experiments; ABA and OK built the robotic platform; OK conducted the experiments and data analysis; All authors contributed to writing the manuscript; OO and MSH supervised the project.

**Competing interests:** The authors declare no conflict of interest.

**Data Availability:** All data needed to evaluate the conclusions in this paper are present in the paper or the Supplementary Materials. The data for this study have been deposited in the Zenodo database (https://doi.org/10.5281/zenodo.21100091). No new materials were generated in this study. No new materials were generated in this work.

Data file S1 to S3